\documentclass[sn-mathphys-num]{sn-jnl}

\usepackage{graphicx}%
\usepackage{multirow}%
\usepackage{amsmath,amssymb,amsfonts}%
\usepackage{amsthm}%
\usepackage{mathrsfs}%
\usepackage[title]{appendix}%
\usepackage{xcolor}%
\usepackage{textcomp}%
\usepackage{manyfoot}%
\usepackage{booktabs}%
\usepackage{algorithmicx}%
\usepackage{algpseudocode}%
\usepackage{algorithm2e}
\usepackage{hyperref}

\begin{document}

\title[Article Title]{SSET: Swapping–Sliding Explanation for Time Series Classifiers in Affect Detection}


\author*[1]{\fnm{Nazanin} \sur{ Fouladgar}}\email{nazanin@cs.umu.se}

\author[2]{\fnm{Marjan} \sur{ Alirezaie}}\email{marjan.alirezaie@oru.se}

\author[1,3]{\fnm{Kary} \sur{ Fr\"amling}}\email{kary.framling@umu.se}

\affil*[1]{\orgdiv{Department of Computing Science}, \orgname{Ume\r{a} University}, \city{Ume\r{a}}, \country{Sweden}}

\affil[2]{\orgdiv{Centre for Applied Autonomous Sensor Systems (AASS)}, \orgname{{\"O}rebro University}, \city{{\"O}rebro}, \country{Sweden}}

\affil[3]{\orgdiv{School of Science}, \orgname{Aalto University}, \city{Espoo}, \country{Finland}}


\abstract{Local explanation of machine learning (ML) models has recently received significant attention due to its ability to reduce ambiguities about why the models make specific decisions. Extensive efforts have been invested to address explainability for different data types, particularly images. However, the work on multivariate time series data is limited. A possible reason is that the conflation of time and other variables in time series data can cause the generated explanations to be incomprehensible to humans. In addition, some efforts on time series fall short of providing accurate explanations as they either ignore a context in the time domain or impose differentiability requirements on the ML models. Such restrictions impede their ability to provide valid explanations in real-world applications and non-differentiable ML settings. In this paper, we propose a swapping--sliding decision explanation for multivariate time series classifiers, called SSET. The proposal consists of swapping and sliding stages, by which salient sub-sequences causing significant drops in the prediction score are presented as explanations. In the former stage, the important variables are detected by swapping the series of interest with close train data from target classes. In the latter stage, the salient observations of these variables are explored by sliding a window over each time step. Additionally, the model measures the importance of different variables over time in a novel way characterized by multiple factors. We leverage SSET on affect detection domain where evaluations are performed on two real-world physiological time series datasets, WESAD and MAHNOB-HCI, and a deep convolutional classifier, CN-Waterfall. This classifier has shown superior performance to prior models to detect human affective states. Comparing SSET with several benchmarks, including LIME, integrated gradients, and Dynamask, we found the superiority of proposed model in terms of explanation quality.}




\maketitle

\section{Introduction}\label{sec.introduction}

Due to the ubiquitous use of machine learning (ML) models, explainable artificial intelligence (XAI)  has become a hot topic in various practical domains \cite{Hardt:2020,Fouladgar-Metrics:2022}. In essence, some ML models, known as “black-box”, have a complex nature which result in reducing transparency, user trust, and debugging capacity. To overcome these challenges, practitioners have paved the way through local (decision) explanations by which feature attributions are provided at a particular instance \cite{Mengnan:2019,Guidotti:2018,Fouladgar-survey:2020}. As an example, let us consider multiple sensory data containing electrocardiogram (ECG), temperature (Temp), and respiration (RESP) measurements collected from an individual in the context of affect recognition. A local explanation method may provide the ECG and RESP measurements contributing to the ML decision of individual's surprised state~\cite{Fouladgar-Conf:2021}. 
 
This example captures explanability for multivariate time series data type. Although our world is surrounded by such data, the literature contains numerous XAI models that focus on images ~\cite{Ghorbani:2019} and tabular data~\cite{guidotti-LORE:2018}. One possible reason for this deficiency is that the evaluation of generated explanations on time series is not as straightforward as the former data for humans. For example, an XAI model makes sense to humans when it gives yellow petals as the prediction explanation of a sunflower~\cite{Ghorbani:2019} or when it gives an age of under 25 as the explanation for rejecting a loan application~\cite{guidotti-LORE:2018}. However, due to the inherent conflation of time and variables, it is difficult for humans to process the significance of each time step in the case of time series data. 

In the literature, two approaches, the \textit{intrinsic} ~\cite{Hsieh:2021,Hosseini:2019,Vinayavekhin:2018} and \textit{post-hoc} ~\cite{Crabbe:2021,Fouladgar-Conf:2021} mechanisms, are usually introduced as local explanations on multivariate time series data. While the former creates built-in modules in the ML design process, focusing on relevant features in terms of attention weights, the latter applies surrogate techniques on top of ML models, assigning attribution scores to the latter features. The study in~\cite{Domjan:2021} shows that providing reliable explanations in terms of attention weights depends on the complexity of datasets, requiring some measures to assess complexity on the lower level of explainability. On the other hand, existing post-hoc approaches fall short of intervening sequences of time steps (context) in the design process \cite{Tonekaboni:2020}. Instead, a step-wise explanation is followed in the time domain. Taking the context into account is crucial in time series applications, specifically affect detection, in which no single time step plays an influential role in the detection of human affective states. Another drawback of post-hoc explanations refers to the inherent differentiability constraint on ML models ~\cite{Crabbe:2021}. Such requirement restricts the applicability of explanations to non-differentiable settings. 

To address these challenges, we propose a post-hoc swapping--sliding decision explanation for multivariate time series classifiers, called SSET. The model includes swapping and sliding stages to detect salient variables and time intervals, respectively. We assign importance scores to the observations in a novel fashion to present explanations. Specifically, in the first stage, we sample train data from target classes in the neighborhood of each instance to be explained. The salient variables are extracted by swapping the instance variables with their counterparts in the train neighbors. In the second stage, we slide a window of selected train data over each time step of the salient variables to thoroughly examine the important sub-sequences.  
The degree of significance is eventually determined by the drop in the black-box prediction score, the window size, and the role of each observation in the sub-sequence construction. Focusing on affect detection, specifically the CN-Waterfall classifier presented in~\cite{Fouladgar_CN:2020}, we show that SSET outperforms three baselines, including Dynamask~\cite{Crabbe:2021}, integrated gradients (IG)~\cite{Sundararajan:2017}, and LIME~\cite{Ribeiro:2016}, on two real-world datasets, WESAD~\cite{Schmidt:2018} and MAHNOB-HCI~\cite{Soleymani:2012}, in terms of \textit{precision} and \textit{informativeness}.


We organize the rest of the paper as follows. In Section~\ref{sec.literature}, we review related works on XAI models tailored to time series data. We present our novel XAI model, SSET, in Section~\ref{sec:Method}, followed by experiments conducted to demonstrate its performance in Section~\ref{sec.Experiments}. Finally, we conclude our study and discuss future research directions in Section~\ref{sec.Conclusion}.

\section{Related Works}
\label{sec.literature}
In this section, we review the literature concerning post-hoc local explanation models tailored to multivariate time series classifiers. As mentioned in Section~\ref{sec.introduction}, in the post-hoc explanations, external models are designed out of the trained ML models to justify the decisions. Studies show the employment of gradient-based approaches in this explanation category \cite{Assaf:2019,Munir:2019,Siddiqui:2020}. In~\cite{Assaf:2019}, two GRAD-CAMs~\cite{Selvaraju:2017} were proposed to measure the output gradients of 2D and 1D convolution neural networks (CNNs) with respect to the networks' feature maps. The goal was to provide spatiotemporal explanations for prediction of the average energy production of photovoltaic power plants and rare server outages. A framework called TSXplain~\cite{Munir:2019} leveraged salient regions of anomalous sequences in an anomaly detection application, using the gradients of output with respect to input components. Similarly, TSInsight \cite{Siddiqui:2020} employs the influence of inputs on the activation of the last layer in a fine-tuned classifier. The model aimed at estimating features importance applicable in the optimization process of the classifier. One of the main drawbacks of the gradient-based approaches is the requirement for the black-box to be differentiable, which limits their applicability in non-differentiable settings. This limitation also applies to the work in~\cite{Crabbe:2021}, where the proposed model, Dynamask, supports mask perturbations. More specifically, Dynamask produces importance scores for each feature and time step by perturbing a mask on the inputs in a gradient-based optimization procedure. Dynamask's strength is that it captures a context for each observation, which was not explicitly addressed in the remaining works discussed here. 

The remaining approaches employ a range of diverse techniques. A two-step procedure, called TSR, was introduced in~\cite{Ismail:2020}. TSR masks each time step to calculate the contribution of observations to the black-box decision. Variables of the salient steps are then occluded to evaluate the variables contribution. The product of the time and variable importance scores represents the final contributions. In \cite{Fouladgar-Conf:2021}, two concepts of contextual importance and utility were employed in an affect detection domain. These concepts rely on perturbing features and measuring output variations to present explanations. In a clinical context ~\cite{Tonekaboni:2020}, temporal shifts in the predictive distribution were quantified to assign the importance scores. Essentially, the model called FIT, defines the scores based on the extent to which a set of features at a time step could best approximate the full outcome distribution. The approximation was facilitated using a recurrent latent variable generator and a KL-divergence operator. Employing a generative adversarial network (GAN) architecture, SPARCE was used ~\cite{Lang:2022} to provide counterfactual explanations. While an LSTM-based generator produced counterfactuals, a discriminator tried to discriminate between the counterfactuals and real samples from the opposite classes. As another counterfactual explanatory model, CoMTE~\cite{Ates:2021} was introduced to generate counterfactuals by means of training neighbors. More specifically, the series of interest were substituted with their neighboring counterparts to evaluate the prediction change. The generated instances which resulted in the highest change and lowest number of substitutions were introduced as counterfactuals.

\section{Method}
\label{sec:Method}
In this section, we begin with a formulation and description of the proposed local explanation for multivariate time series classifiers, namely SSET\@. 

\subsection{Preliminaries}
\label{sec.preliminaries}
Let $X_{tr} = \{ x^\prime \in \mathbb{R}^{T \times V} \}$ and $X_{te} = \{ x \in \mathbb{R}^{T \times V} \}$ be sets of train and test data, respectively, where $\mathbb{R}$ indicates real numbers in the range $[0, 1]$. Moreover, $T$ and $V$ are the number of time steps and variables (signals), respectively. We assume the trained black-box $f$ produces probability scores $y = f(x)$ for each test instance $x$ over a set of classes $Cl = \left\{1,2,\dots, C\right\}$, where $C$ denotes the total number of classes. This set includes a winner class $c$. Throughout the remainder of the paper, the  following notation will be used:

\begin{itemize}
 \item $\overline{Cl}$: a set of all classes excluding $c$ (or target classes)
 \item $s \in S = \left\{1,\dots, V\right\}$: a signal from a set of signals $S$
 \item $t \in \left\{1,\dots, T\right\}$: a time step 
  \item $x_i$: the test instance of interest
  \item $y_{i}^c$: the prediction score of $x_i$ over the class $c$ 
 \item $x^{1:T, s}$: values of $x$ in the signal $s$
  \item $x^{t-j:t+k, s}_i$: values of a sub-sequence between $[t-j, t+k]$ time steps in the signal $s$ of the instance $x_i$, where $j, k \in \left\{0,\dots, T-1\right\}$
 
 \end{itemize}

\subsection{Proposed XAI Model: SSET}
\begin{figure*}[t]
		\begin{center}
			\centering
			\includegraphics[scale=0.4]{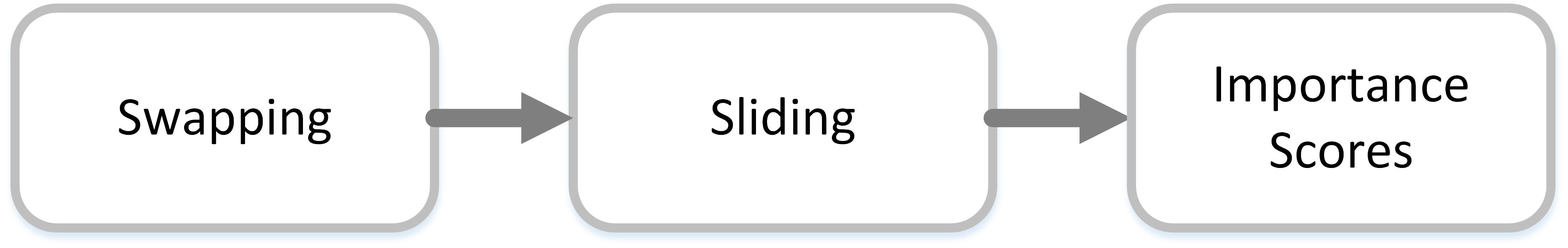}\\
			\caption{SSET schema showing the three main components.}
			\label{fig.Schema}
		\end{center}	
\end{figure*}

We present our proposed local explanation model, SSET, with a focus on the affect detection domain. In this field, the data are usually collected from physiological sensors~\cite{Schmidt:2018,Soleymani:2012}, which are used to record a cascade of physiological processes occurring when humans are subject to different stimuli~\cite{Giannakakis:2019}. Taylor~\cite{Taylor:1959} argues that these processes have a duration of less than $90$ seconds. The argument could imply that for series sampled at more than $1$~Hz, manipulating only one time step may have no effect on detecting a human's mental state. In other words, to drive the decision explanation of the detector, we need to consider the manipulation of several steps and then track the outcome variations. In this paper, we define a \textit{context} or neighboring steps for each observation to scrutinize the variations and detect salient sub-sequences. However, to alleviate the cognitive burden on end users when multiple sensory data are employed, we focus on salient sub-sequences of important physiological sensors. To specify the degree of salience, we further present a novel formulation considering the impact of the context size, among other criteria. The procedure 
yields to meaningful explanations for the detection of human affective states. Algorithm~\ref{algorithm} shows the SSET process. In the following, we describe the three main components of the model (Fig.~\ref{fig.Schema}), including \textit{swapping}, \textit{sliding}, and calculating \textit{importance scores} in detail. 

\subsubsection{Swapping} 
\begin{figure*}[ht]
		\begin{center}
			\centering
			\includegraphics[scale=0.48]{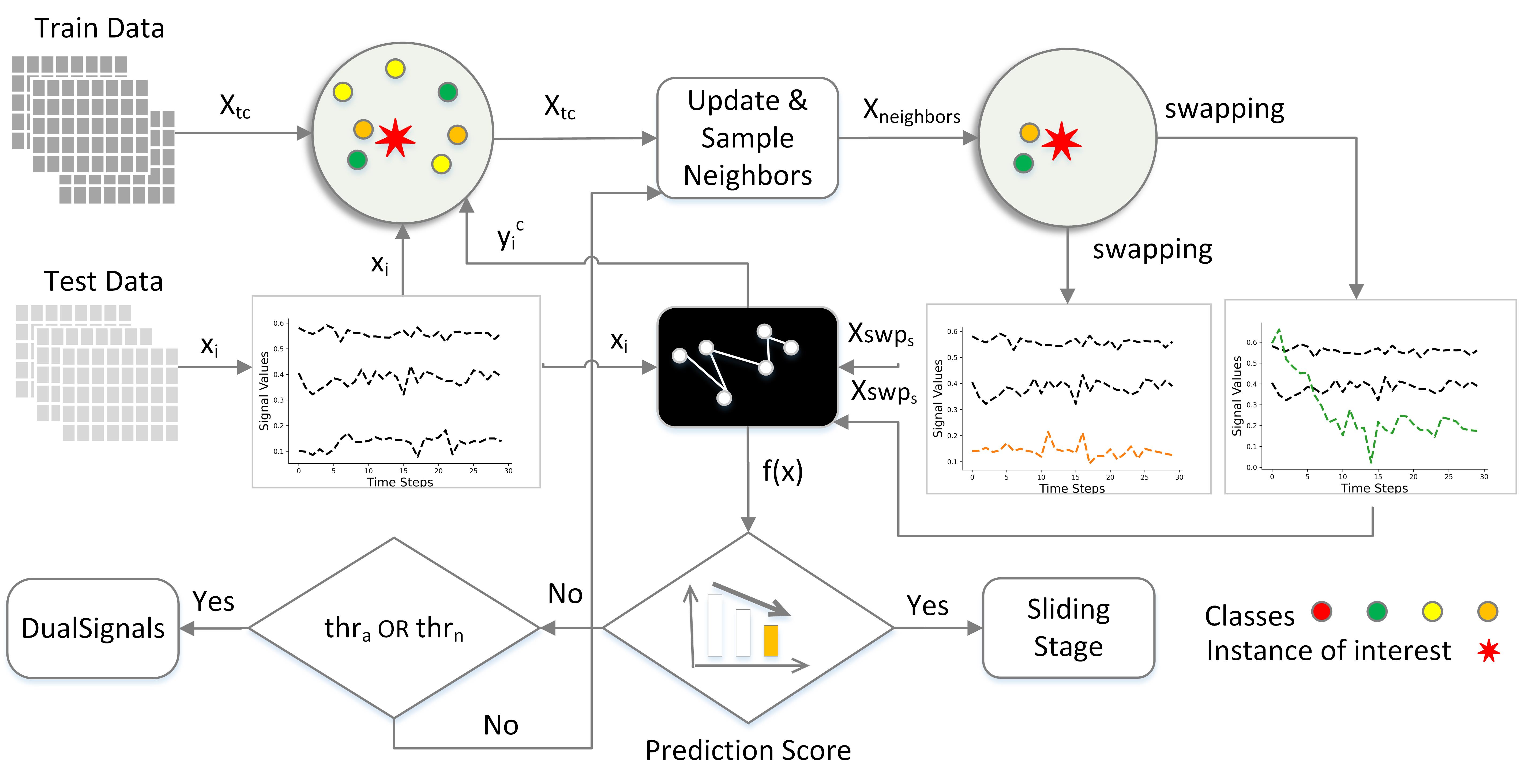}\\
			\caption{Swapping stage of SSET, providing salient signals. In this stage, the black-box predicts class $c$ (shown by the red color) with the score $y_i^c$ for the series of interest $x_i$ (red star). We extract the train data with opposite classes to $c$ (colored circles), denoted as $X_{tc}$, from which a few neighbors $X_{neighbors}$ are randomly sampled (see Fig.~\ref{fig.neighborhood}). To detect the important variables, the values of each signal $s$ in $x_i$ is swapped with its counterparts in $X_{neighbors}$, resulting in $X_{swp_s}$. The corresponding signals of those $X_{swp_s}$ that cause a performance drop maximally and below a certain threshold (orange series), are considered as salient. Otherwise, either further sampling and updating of the neighborhood scope are performed, or the \textit{DualSignals} component is activated.}

			\label{fig.Swapping}
		\end{center}	
\end{figure*}

\begin{figure*}[t]
		\begin{center}
			\centering
			\includegraphics[scale=0.35]{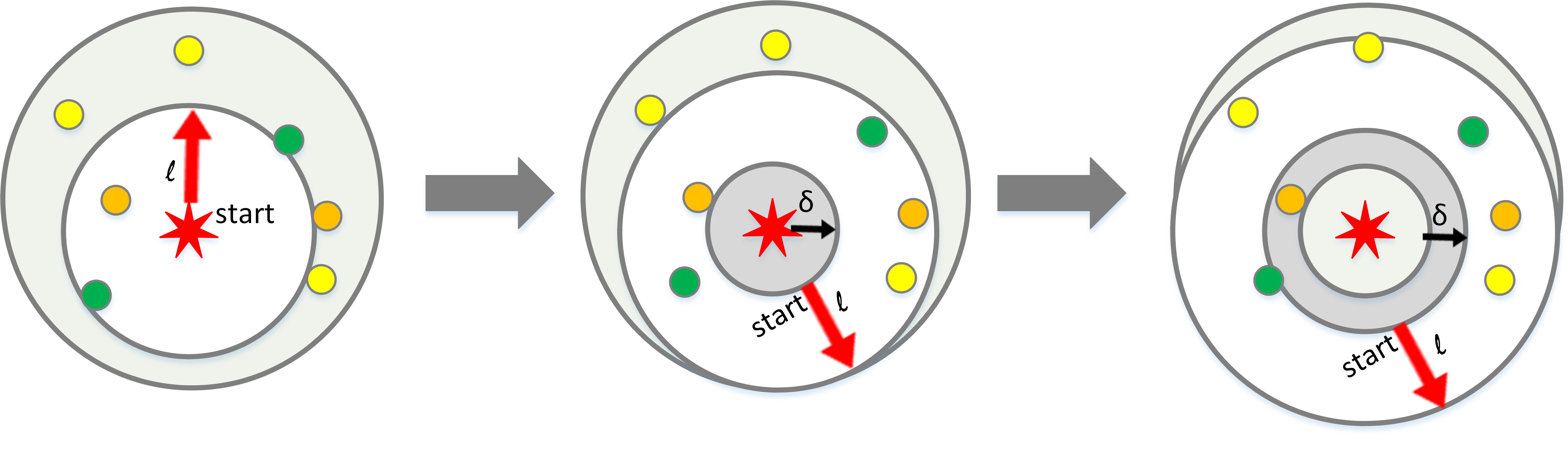}\\
			\caption{Determining the neighboring scope. Starting from the instance to be explained (the red star), a region of radius $l$ (white space) is specified around the instance. If no neighboring samples (colored circles) fall in this space or satisfy the dropping constraint, the scope is updated by shifting the start point by $\delta$. }
			\label{fig.neighborhood}
		\end{center}	
\end{figure*}

In this stage, we wish to distinguish the salient physiological signals ($V_{imp}$) contributing to the detection of a human's mental state. Focusing on these signals helps the next stage to produce informative explanations. As mentioned in Algorithm~\ref{algorithm} and shown in Fig.~\ref{fig.Swapping}, first, a set of random instances is selected from the test data. For each instance $x_i \in X_{te}$ (the red star in Fig.~\ref{fig.Swapping}), the black-box model predicts $y_{i}^c$. The train data with target classes in $\overline{Cl}$ (the colored circles in Fig.~\ref{fig.Swapping}) are then extracted from the dataset, denoted as $X_{tc} \subset X_{tr}$. We replace each signal in $x_i$ with its counterparts in $X_{tc}$, while keeping other signals unchanged. In other words, the values of $x_{i}^{1:T, s}$ are manipulated by those in $X_{tc}$. Using the train data in this process guarantees that the generated instances fall in an in-distribution space.
However, as there can be large numbers of $X_{tc}$ instances in large datasets, providing the explanation will be computationally expensive. Therefore, we randomly sample a user-defined number of neighbors $X_{neighbors} \subset X_{tc}$ by which the replacement of signal values is accomplished. Fig.~\ref{fig.neighborhood} illustrates how the neighborhood scope is defined, and we will also explain it in more detail.

We denote the swapped instances corresponding to the signal $s$ by $X_{swp_s}$. Evaluating the instances and using in the next stage, we select those which cause maximum drop in the black-box performance below a certain threshold, $thr_{c}$. The goal is to increase the chance of changing the winner class to a target class in $\overline{Cl}$. As a result, the respective signals are appointed salient and shown by $S_{imp}$: 

\begin{equation}
\label{eq.important_sig}
S_{imp} = \{s\in S| \max \{f(X_{swp_s}) \leq thr_c\} \} 
\end{equation}

In the above procedure, one can infer that the train neighbors play a vital role in identifying the salient signals. Here, we explain how the neighborhood scope is defined and what strategies SSET takes to extract the neighbors. We first consider the region within a distance $l$ of $x_i$ (the white circle in Fig.~\ref{fig.neighborhood}), where a number of train instances is sampled randomly from the target classes. In case the instances are distributed outside the region or no promising observations are detected in the region, the sampling is repeated until a certain number of attempts ($thr_a$) is met. By promising observations, we mean the neighbors which cause the prediction drops below the threshold $thr_c$. This strategy improves the chance of finding promising neighbors. We further update the neighborhood region and reset the attempts to extend the explorations, provided that the sampling failures. The region is updated by shifting the exploration space to $\delta$. Formally, we are interested on any promising neighbor $x_{n} \in X_{neighbors}$ whose Euclidean distance to the instance of interest falls in this region:

\begin{equation}
\label{eq.neighbors_scope}
start + \delta \leq \lVert x_i - x_{n} \rVert \leq start + \delta + l
\end{equation}
where:
    \begin{align*}
       start + \delta + l  \leq thr_n
    \end{align*}
In the Equation above, we restrict the neighborhood scope and the updates by a threshold $thr_n$.

Given the swapping stage and mentioned strategies, according to our observations (see Section~\ref{sec.Experiments} for details), only one signal contributes to the black-box decision in most test data. For the data without any individual salient signals, a \textit{dual-signals} swapping mechanism (shown as $DualSignals$ in Fig.~\ref{fig.Swapping} and Algorithm~\ref{algorithm}) is performed. In this mechanism, we take inspiration from the CN-Waterfall architecture ~\cite{Fouladgar_CN:2020}, in which correlated and non-correlated signals are distinguished. We extract the respective sets of signals, and perform a similar swapping procedure as before, yet on each set and pair of signals.

\subsubsection{Sliding}
\begin{figure*}[ht]
		\begin{center}
			\centering
			\includegraphics[scale=0.53]{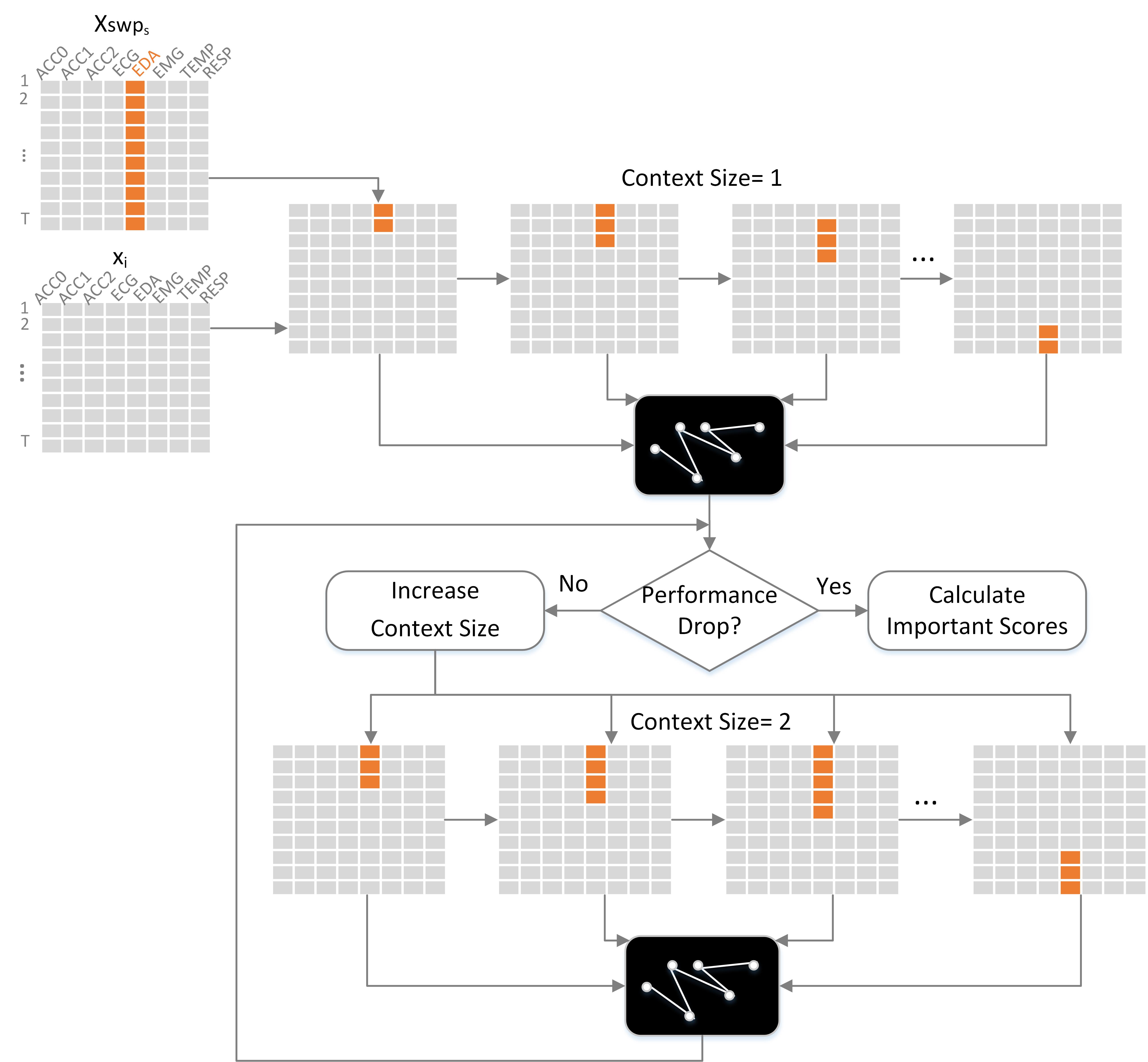}\\
			\caption{Sliding stage of SSET, providing salient sub-sequences of the important signals. To simplify the visualization, we only show one salient signal (EDA) in $x_i$ using the orange color. Starting with a context (neighboring step) size of $1$, a window is slid over EDA, including the current, backward, and forward time steps. The corresponding values are replaced by their counterparts in $X_{swp_s}$, while the values of other time steps in $x_i$ are kept unchanged. Provided that the performance of the black-box drops in the manipulated instances, the corresponding sub-sequences are designated as salient, and their importance degree is measured for each time step. Otherwise, the context size is increased and the process is repeated iteratively.}

			\label{fig.Sliding}
		\end{center}	
\end{figure*}

The goal of this stage is to provide important sub-sequences of the signals selected by the swapping stage and present them as the contributors to the black-box decision. To achieve the goal, we consider a context ($ctx$) with an adaptive size for each observation of the salient signals (Fig.~\ref{fig.Sliding}). The context stands for the neighbors of each time step constructing sub-sequences in the signals (see Section~\ref{sec.preliminaries}). Sliding a window $w$ of the same size as the sub-sequences, the corresponding values in $x_i$ are replaced by their counterparts in $X_{swp_s}$, while the remaining observations are unchanged. This way, $T$ manipulated instances ($X_{sld}$), are generated in each salient signal and for a specific window size. To find the salient sub-sequences, we evaluate the performance loss of the manipulated instances at the winner class. If the performance drops below $thr_c$, the respective context and time step are counted important. We show these sub-sequences as $Seq_{imp}$ and formulate in the following:
\begin{equation}
\label{eq.important_seq}
Seq_{imp} = \{x^{t-j: t+k, s}_i \in x_{sld_t} | f(x_{sld_t}) \leq thr_c\} 
\end{equation}
where:
\begin{align*}
    x_{sld_t} \in X_{sld}, s \in S_{imp}
\end{align*}
Here, $x_{sld_t}$ refers to the respective manipulated instance at time step $t$.

We further enrich the context by increasing the number of neighbors and correspondingly increase the window size, in case the former context does not play an influential role in the output. The procedure continues until at least one sub-sequence satisfies the dropping constraint and is determined to be salient. 

In general, the sliding process only applies when signals independently contribute to the black-box decision. We extend the process when the \textit{dual-signals} functionality is activated. As previously discussed, this functionality specifies the contribution of signal pairs simultaneously. To inspect the salient sub-sequences, a common context size is defined for each observation in the salient pair over which the constructed window slides. Like before, we adaptively increase the context size and iteratively repeat the process until salient sub-sequences are detected. 

\subsubsection{Importance Score}
In this stage, we quantify the importance score of salient sub-sequences and present the explanation of the instance to be explained. In Equation.~\ref{eq.imp_score_timestep}, we formulate the scores in a novel fashion, incorporating multiple factors: the drop in the black-box output, the window size, and the role of each time step i.e., either “current” or “neighbor”.

As mentioned before, the salient sub-sequences are taken to be the ones by which the black-box output drops. 
The larger the drop is, the higher the importance of the sub-sequence will be. 
However, relying solely on the amount of performance loss may not accurately reflect the importance of sub-sequences with different sizes, yet the same amount of loss. To address this limitation, we take the impact of the window size into account in each time step. Recalling the informativeness of the explanation from the sliding stage and to reduce the cognitive burden on users, smaller salient sub-sequences are expected to be more informative than larger ones. Therefore, we incorporate an exponential effect of window size $\lvert w \rvert$, with respect to the total number of time steps. To avoid producing scores greater than $1$, we further regulate the contribution of the window size by a coefficient $\lambda$ and impose a minimum operator on the incorporated factors. As any salient sub-sequence is the result of sliding $w$ over each time step, we distinguish the role of steps. The time step $t^\prime$ over which the window is constructed takes the role of “current”, while other steps in the window are considered as “neighbor”. It is assumed that the current step is the source of output drop, thereby a higher importance score is assigned to this step than its neighbors. We resort to an $\alpha$ proportion of the score in the current step, as the importance of its neighbors. In the following, we address the process formally:

\begin{equation}
\label{eq.imp_score_timestep}
\underset{t \in Seq_{imp}}{I_{t}}  = 
\begin{cases}
    \min \left(\lvert y_i^c - y_m^c \rvert + \lambda \times \exp(-\frac{\lvert w \rvert}{T}), 1\right) & \text{if $t = t^\prime $} \\
    \alpha \times I_{t^\prime} & \text{Otherwise}
    
\end{cases}
\end{equation}
where $I_{t}$ is the important score at time $t$ of the salient sub-sequence. The prediction score of the manipulated instance at the winner class $c$ is also denoted as $y_m^c$.

Although the Equation~\eqref{eq.imp_score_timestep} can capture the importance score of each time step in the salient sub-sequence, it doesn't hold under overlapping salient sub-sequences. In this case, the observations could take both the “current” and “neighbor” roles resulting in multiple importance scores. To tackle the problem, we accept the highest score in each time step. Note that the steps and signals other than those in $Seq_{imp}$, respectively, are assigned zero importance. We formulate the final scores in a matrix, namely $IMP$: 

\begin{equation}
\label{eq.imp_score}
IMP =
\begin{cases}

   {\max} \left(I_{t}\right) & \text{$t \in Seq_{imp}$}\\
   $0 $& \text{Otherwise}
\end{cases}
\end{equation}

It is worth mentioning that a similar regime of measuring the importance scores is imposed for pairs of salient sub-sequences when the \textit{dual-signal} functionality is activated.

\begin{algorithm}
\caption{SSET Explanation Procedure.}\label{algorithm}
\KwData{$X_{te}, X_{tr}, f$}
\KwResult{$IMP$}
$start = -1$\\
\For{$x_i \in Select(X_{te})$}{
    $y_{i}^c \gets f(x_i)$\\
    $X_{tc} \gets Select (X_{tr})$\\    
    \While{$!S_{imp}$} {
        $a = 0$\\
        $start += \delta$ \\
        $n = start + l$\\
        \If{$n \leq thr_{n}$}{
            \While{$a \leq  thr_{a}$}{
                $a++$\\
                $X_{neighbors} \gets SampleNearest (x_i, X_{tc})$\\
                \If {$X_{neighbors}$}{
                    $X_{swp_s}, S_{imp} \gets Swap (x_i, X_{neighbors})$
                    \\
                    \If {$S_{imp}$}{
                        $ Seq_{imp} \gets  Slide (x_i, X_{swp_s}, S_{imp})$\\    
                        $IMP \gets ImpScores (x_i, Seq_{imp})$\\
                        $\textbf{break}$
                    }
                } 
            }
        }{
          $DualSignals(x_i)$\\
          $\textbf{break}$ 
        }
    }
}
\end{algorithm}

\section{Experimental Results}
\label{sec.Experiments}

In this section, we evaluate the quality of SSET and compare it with benchmarks. 

\subsection{Datasets} We employed a public and an academically available dataset, WESAD and MAHNOB-HCI, respectively. The datasets were preprocessed as discussed in~\cite{Fouladgar_CN:2020}. In short, the signals of eight sensors were extracted from the WESAD dataset, including three-axis accelerometer (ACC0, ACC1, ACC2), electrodermal activity (EDA), electromyogram (EMG), RESP, ECG, and TEMP data. We selected four affective states---\textit{neutral}, \textit{amusement}, \textit{stress}, and \textit{meditation}---from the data of all participants, resulting in $433\,350$ samples. The data of the participants were then unified in terms of length, downsampled to $10$~Hz, normalized to the range $[0, 1]$, and segmented into series of 3-second windows (30 time steps) with 1-second overlaps (10 time steps). A similar procedure was applied to the MAHNOB-HCI dataset but on the signal data from seven sensors, including three-channel ECG (ECG1, ECG2, ECG3), two-channel galvanic skin response (GSR1, GSR2), RESP, and TEMP data. Three affective states---\textit{amusement}, \textit{happiness}, and \textit{surprise}---with $13\,440$ instances were examined for the latter dataset.

\subsection{Black-box} We used CN-Waterfall, a deep convolutional affect recognizer, which has been shown to be superior to other traditional and deep learning models~\cite{Fouladgar_CN:2020}. Briefly, this black-box consists of  \textit{Base} and \textit{General} components, providing different levels of data representation. It is worth mentioning that in the \textit{General} component, correlated and non-correlated data representations are fused independently to extract signals interrelation information. Moreover, CN-Waterfall was trained on data randomly split into train and test sets in an $80:20$ ratio.

\subsection{Benchmarks} To evaluate the explanations produced by SSET, we included local explainers from different categories: IG and LIME as gradient- and perturbation-based approaches, respectively, which have been widely applied to non-time-series data types, and Dynamask as a perturbation-based approach that has mainly been applied to time series datasets.

\subsection{Settings} In SSET, different user-defined hyperparameters has been introduced. We select $200$ test and $10$ train data in the neighborhood of the instance to be explained from each dataset. The neighborhood scope is initialized as the space between $0$ and $l = 1$ and shifted by $\delta = 0.1$ when needed. Therefore, we set $start = -1$ to meet the space constraint. We also adjust the maximum neighborhood after shifting the scope to $thr_n = 8$, empirically. In addition, finding the neighboring samples is attempted for a maximum of $thr_a = 10$ times. An intuitive choice of threshold, $thr_c = 0.5$, is considered as the measure of prediction drop. However, any desired value could be initialized. We consider the context $ctx = 1$ to produce as informative explanation as possible. Moreover, $\lambda = 0.1$ is initialized empirically. To avoid a significant suppression in the neighbors contribution, we select $\alpha = 0.9$.

In IG, the baseline was taken to be the average of the train data. The number of steps along a path from the baseline was $10$, at which the gradients were calculated. Regarding LIME, we limited the number of generated samples to $50$, by which LIME approximates a linear model. In Dynamask, the same configurations as those in the original paper~\cite{Crabbe:2021} were examined using the deletion variant and  fade-to-moving average perturbation $\pi^m$. 

\subsection{Explanations Evaluation} 
\label{sec.explaination-comparison}
\begin{figure*}[ht]
    \centering
    \includegraphics[width=0.3\textwidth]{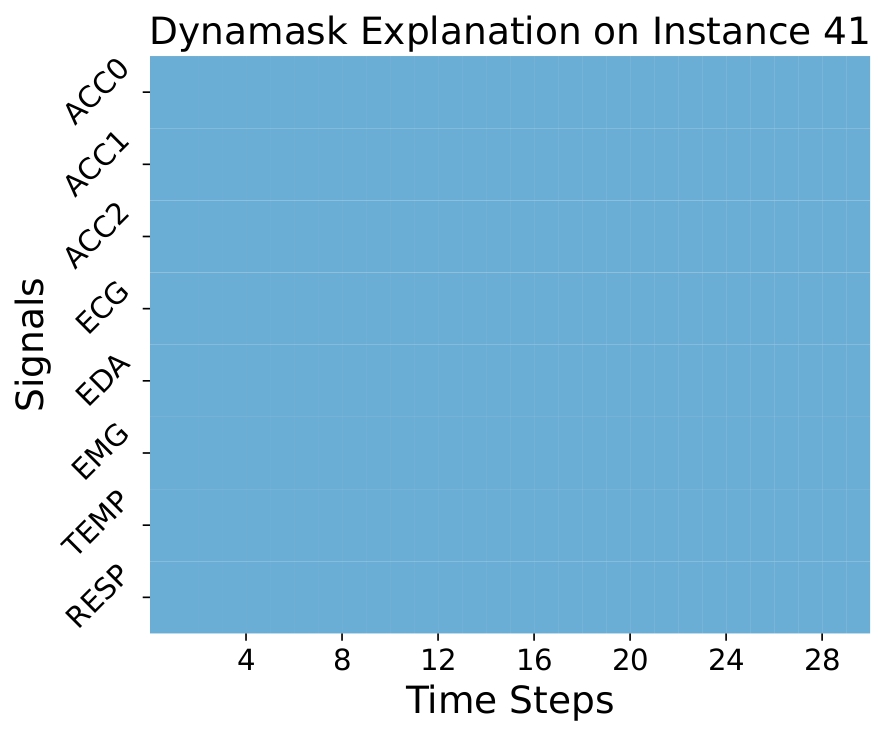} 
    \includegraphics[width=0.3\textwidth]{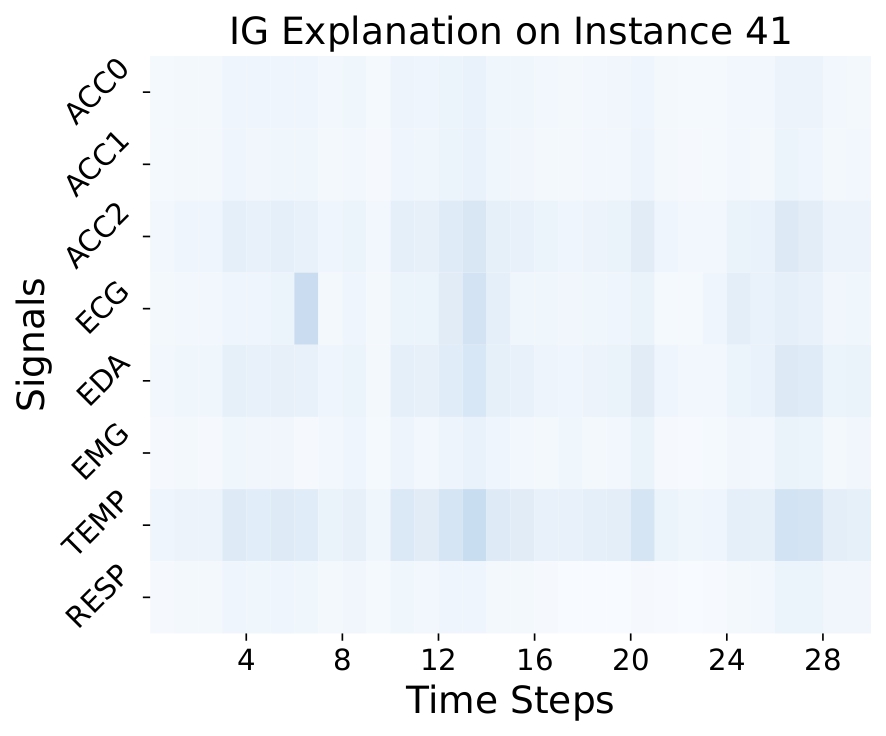}
    \includegraphics[width=0.3\textwidth]{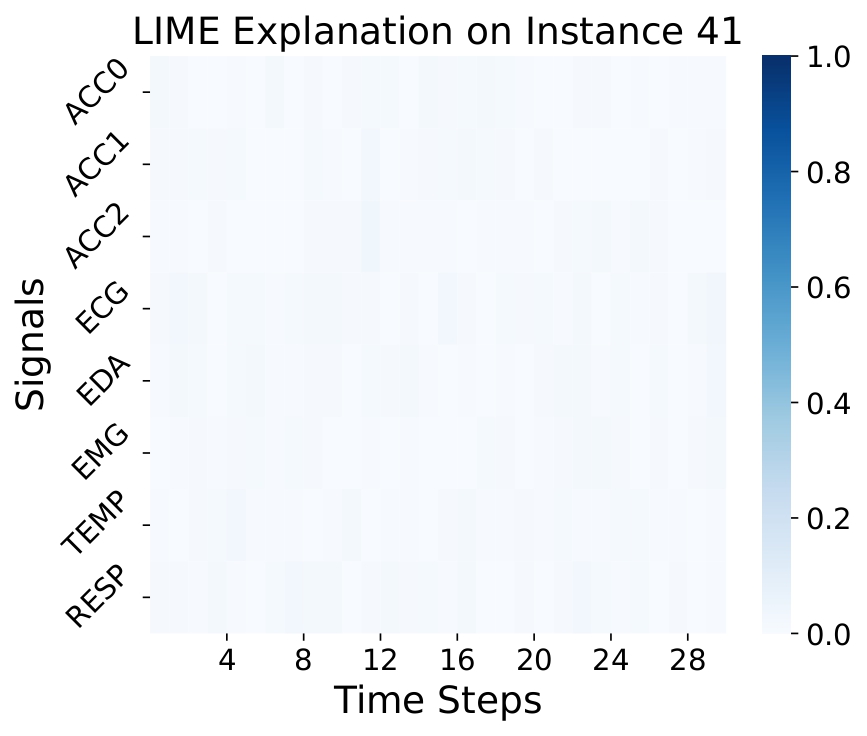}
    \includegraphics[width=0.3\textwidth]{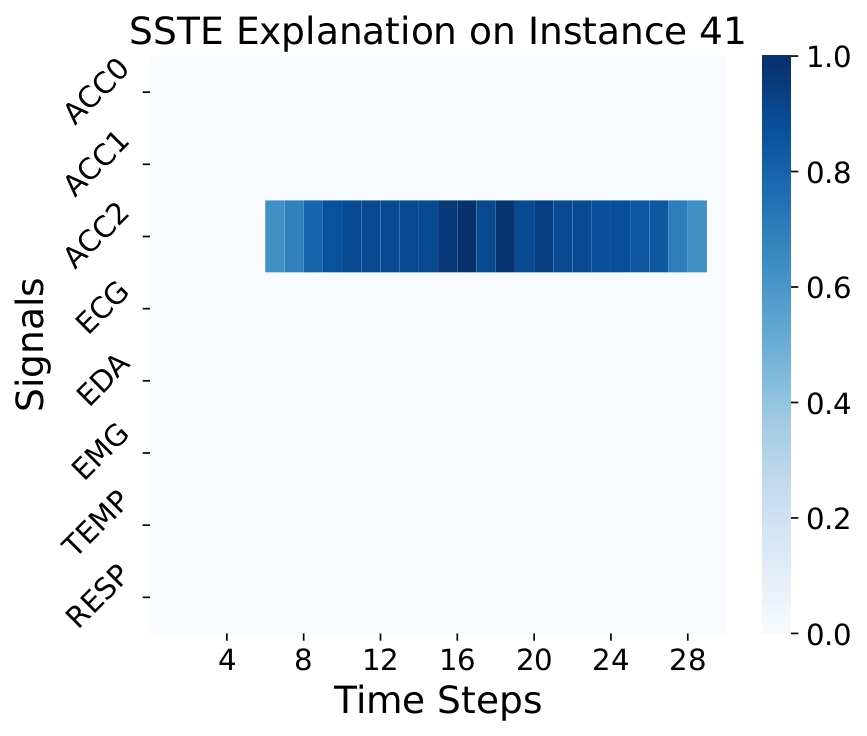}
   
    \caption{Dynamask (a), IG (b), LIME (c), and SSET (d) explanations of instance $41$ on WESAD\@. The explanations are in the form of importance scores, ranging between 0 and 1 and mapped to blue colors in the heatmap visualization. A darker color indicates higher importance. As can be seen, none of the models except SSET provide a meaningful explanation. The Dynamask scores are locked at $0.5$ for all features, IG assigns low scores to most features, and LIME concludes that nearly all features are unimportant. However, SSET presents the sub-sequence of ACC2 between time steps 7 and 29 as highly important in the detection of the \textit{neutral} state, aligning with the domain knowledge.}
    \label{fig.explaination_WESAD}
\end{figure*}

In this subsection, we investigate how SSET generates explanations compared to the benchmarks on the WESAD and MAHNOB-HCI datasets. Fig.~\ref{fig.explaination_WESAD} shows a heatmap visualization of explanations for the instance $41$ in WESAD\@. The vertical and horizontal axes show the eight signals and $30$ time steps of the instance, respectively. The color bar highlights the importance scores for each variable and time step. The darker the color, the closer the scores are to $1$ and the more important the elements are.

As observed, none of the benchmarks provide informative explanations. Dynamask (Fig.~\ref{fig.explaination_WESAD}(a)) is mainly locked to importance scores equal to $0.5$, the initial value of the mask, for all the time steps and variables. This is because Dynamask requires the black-box model to be differentiable with respect to the perturbed data. As the perturbed data result from applying the mask to the instance to be explained and the average values of its variables, there is insufficient variation in the affective states. Therefore, the black-box model of our application, CN-Waterfall, produces gradients close to $0$ with respect to the perturbed data. In other words, the mask (the importance scores of the instance to be explained) does not change even after several rounds of performing Dynamask. These considerations also apply to the other selected test instances. 

Regarding the IG explainer (Fig.~\ref{fig.explaination_WESAD}(b)), we can see that not only are the scores relatively close to $0$, but they are also distributed across all variables and steps. IG uses the gradients of the black-box output with respect to the instances in a straight-line path. If the black-boxes, such as CN-Waterfall, show poor gradients on the instances, the average gradient inclines to $0$. In addition, IG uses element-wise calculations of the gradients, resulting in the non-contextual distributed scores.

LIME (Fig.~\ref{fig.explaination_WESAD}(c)), like IG, does not provide a valid explanation, and the importance scores are nearly $0$.
This model generates several perturbed data points around the instance of interest and then learns a linear approximation model. Due to the entangled classes in the affective computing domain, the approximated linear model cannot distinguish the nonlinear class boundaries. LIME also suffers from requiring a context-wise realization of time steps to generate the importance scores.

We argue that SSET highlights the most informative part of the series as the explanation of the instance. In Fig.~\ref{fig.explaination_WESAD}(d), we observe that SSET chooses only the salient sub-sequence rather than a distributed or monotonic explanation at the instance $41$. Here, the time steps between 10 and 27 show the importance scores are higher than $0.8$, while the steps between 7 and 9 as well as 28 and 29 contribute roughly $0.7$ to the CN-Waterfall state of \textit{neutral}. Other time steps and signals have no impact on this decision. These results agree with domain knowledge. Since the participants were asked to sit or stand while data were collected for this state, the accelerometer signal could impact the discrimination of the \textit{neutral} state for a specific participant. Furthermore, the sub-sequence of time steps has been correctly presented, owing to a $0.8$ drop in the output score when performing the sliding stage.

\begin{figure*}[t]
    \centering
    \includegraphics[width=0.3\textwidth]{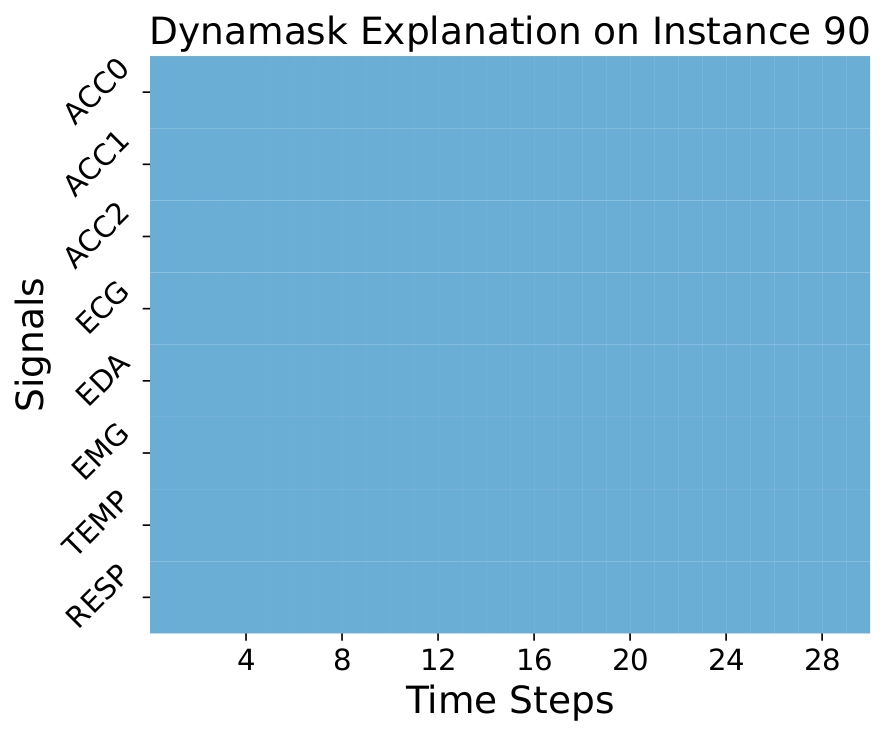} 
    \includegraphics[width=0.3\textwidth]{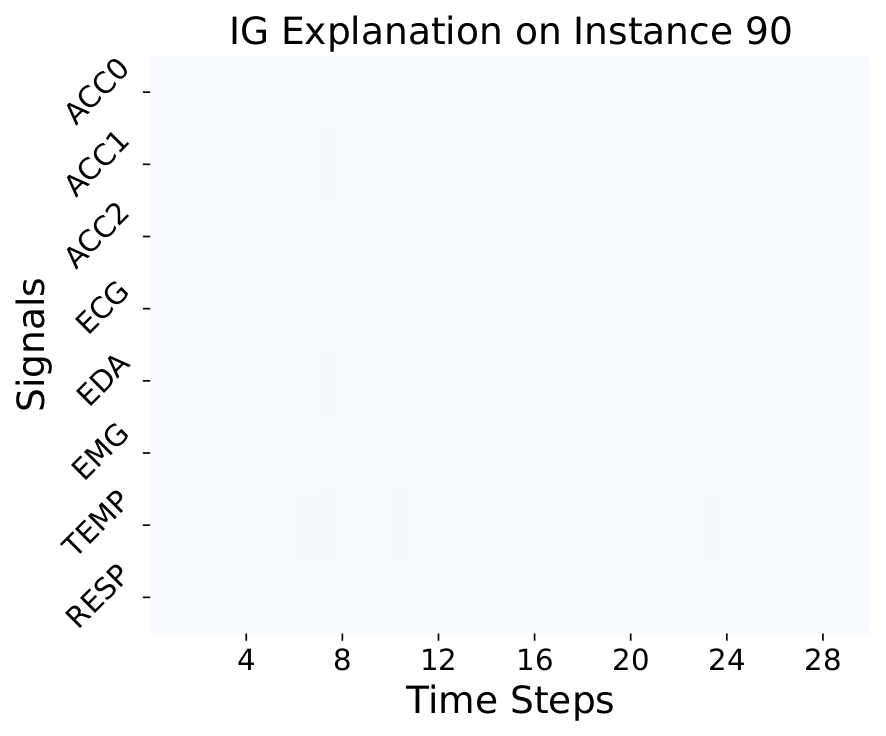} 
    \includegraphics[width=0.3\textwidth]{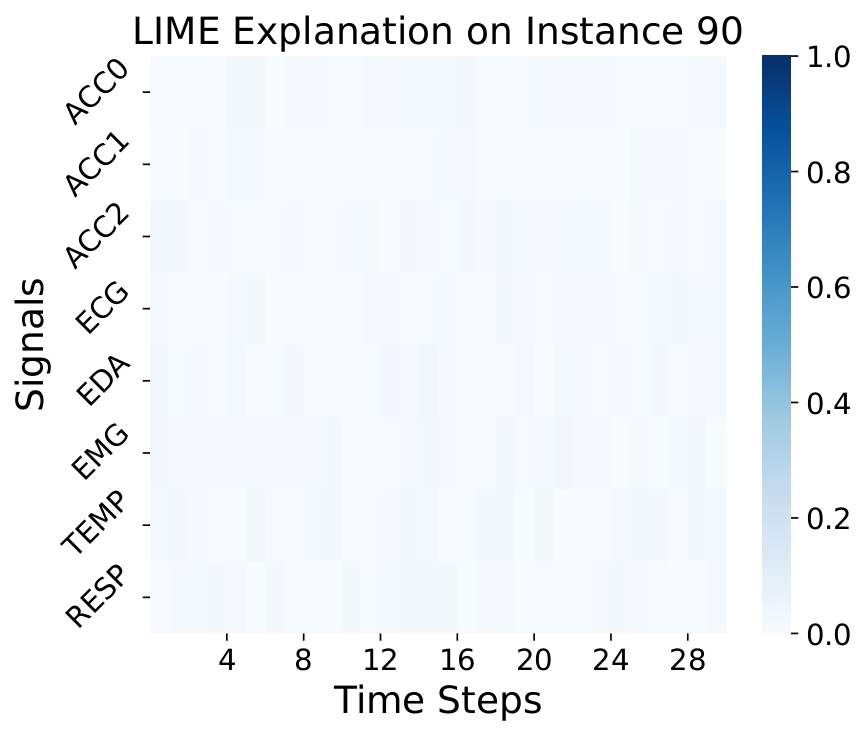}
    \includegraphics[width=0.3\textwidth]{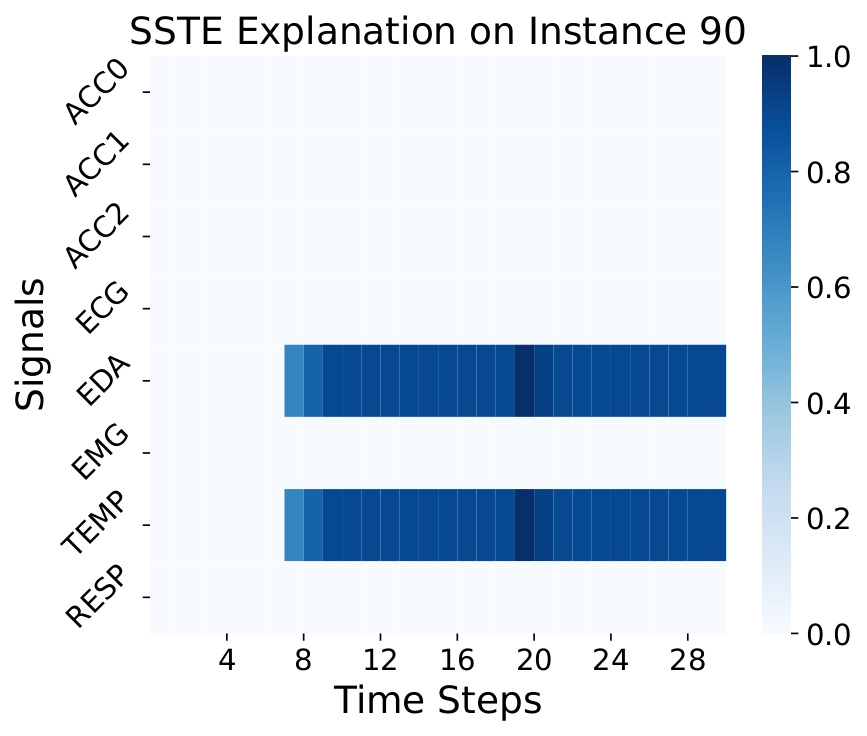}
    \caption{Dynamask (a), IG (b), LIME (c), and SSET (d) explanations of instance $90$ in WESAD, when the \textit{dual-signal} functionality is activated. Similar to Fig.~\ref{fig.explaination_WESAD}, explanations are provided by the heatmap visualization with the color map for the importance scores in the range 0--1. We can infer that SSET provides a meaningful explanation for the \textit{neutral} decision by detecting the sub-sequences of EDA and TEMP\@. The proposed model simultaneously assigns a score of $0.9$ to the steps between 10 and 30, and $0.7$ to steps $7$ and $8$, for both signals. The benchmarks fail in this case, consistently showing scores of either $0.5$ or $0$.}
    \label{fig.explaination_dual_WESAD}
\end{figure*}

We also compared the SSET explanation with the benchmarks on WESAD when one signal alone cannot influence the decision and so the \textit{dual-signal} functionality is activated. In Fig.~\ref{fig.explaination_dual_WESAD}, a heatmap visualization of explanations is presented for instance $90$. As with the earlier case, the benchmarks fail to produce valid and informative explanations, while SSET highlights two sub-sequences of the salient signals in the decision of the \textit{neutral} state. As human physiology differs between subjects~\cite{Schmidt:2019}, one may expect different reactions to the same state. Here, the reactions are associated with the EDA and TEMP data, which jointly contribute to the detection of the \textit{neutral} state. In these signals, the time steps from $10$ to $30$ show importance above $0.9$ with the highest in step $20$. Steps $8$ and $9$ have an importance of $0.7$ on average. Other steps and signals do not affect the state in this instance.  

\begin{figure*}[t]
    \centering
    \includegraphics[width=0.3\textwidth]{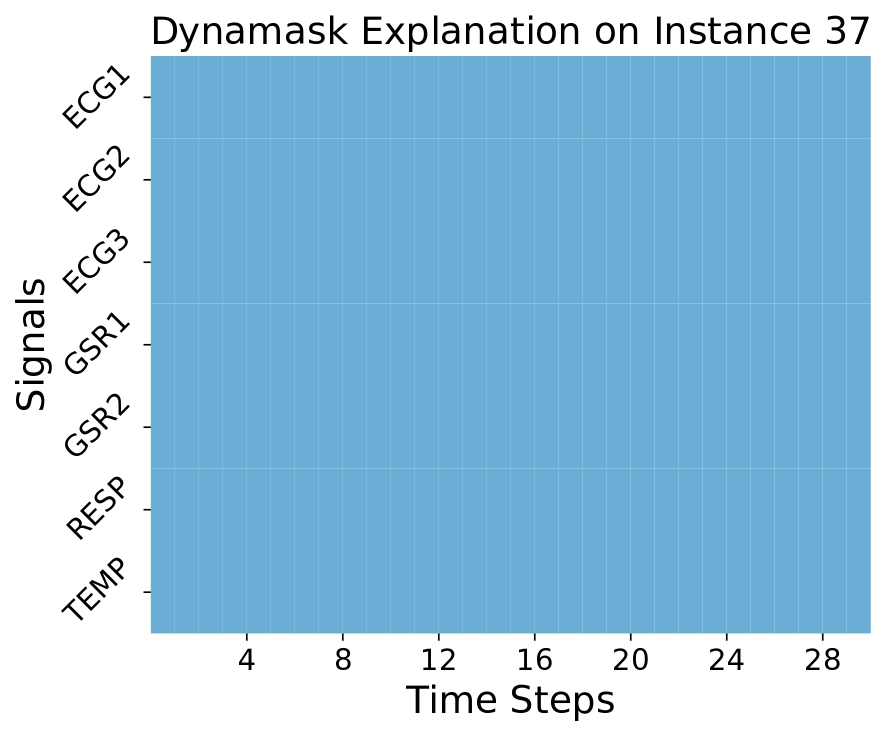}
    \includegraphics[width=0.3\textwidth]{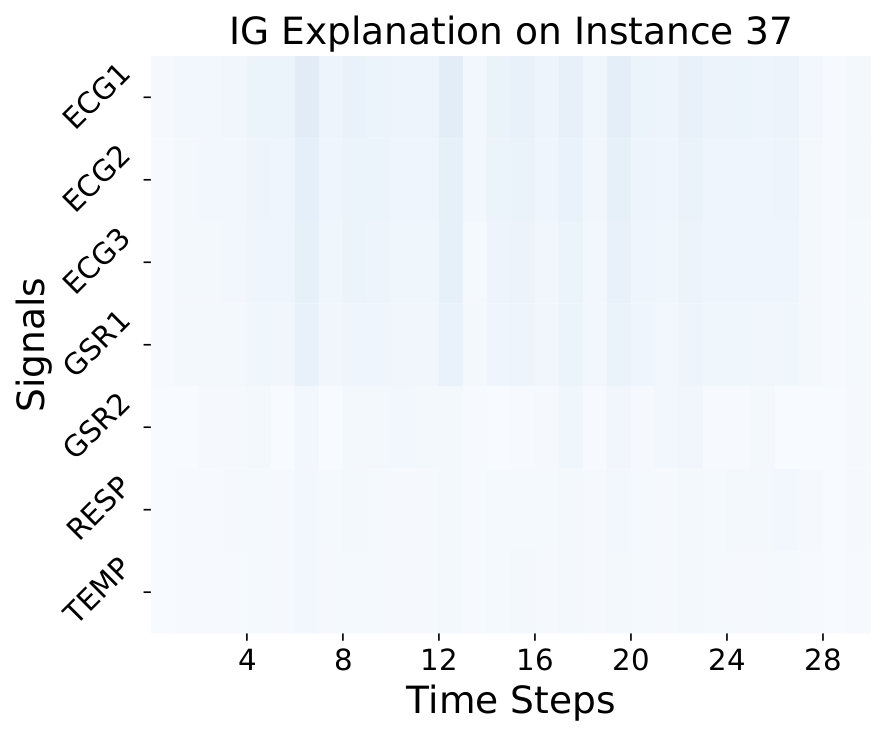} 
    \includegraphics[width=0.3\textwidth]{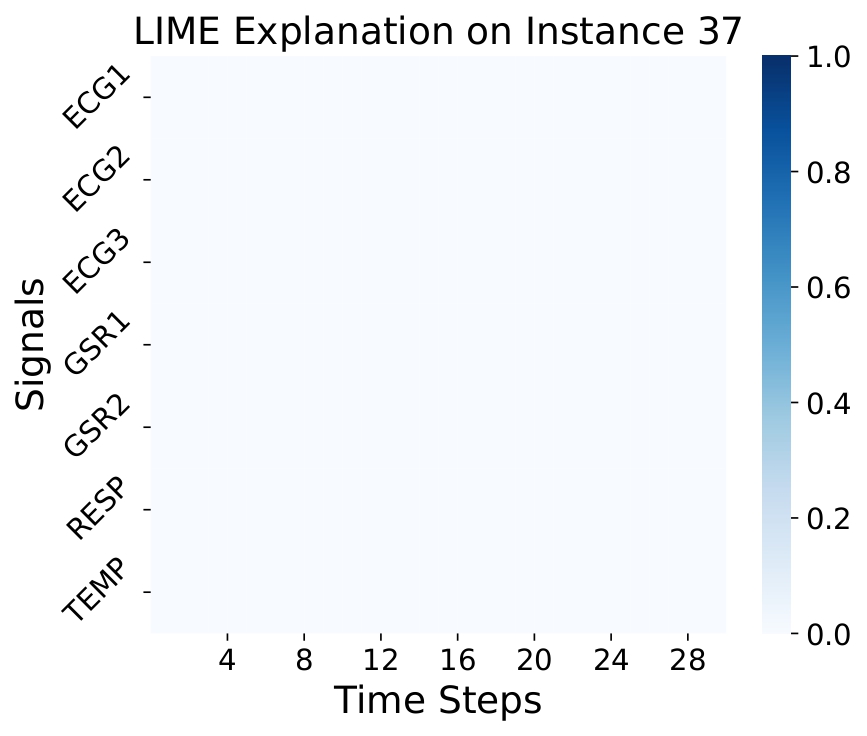}
    \includegraphics[width=0.3\textwidth]{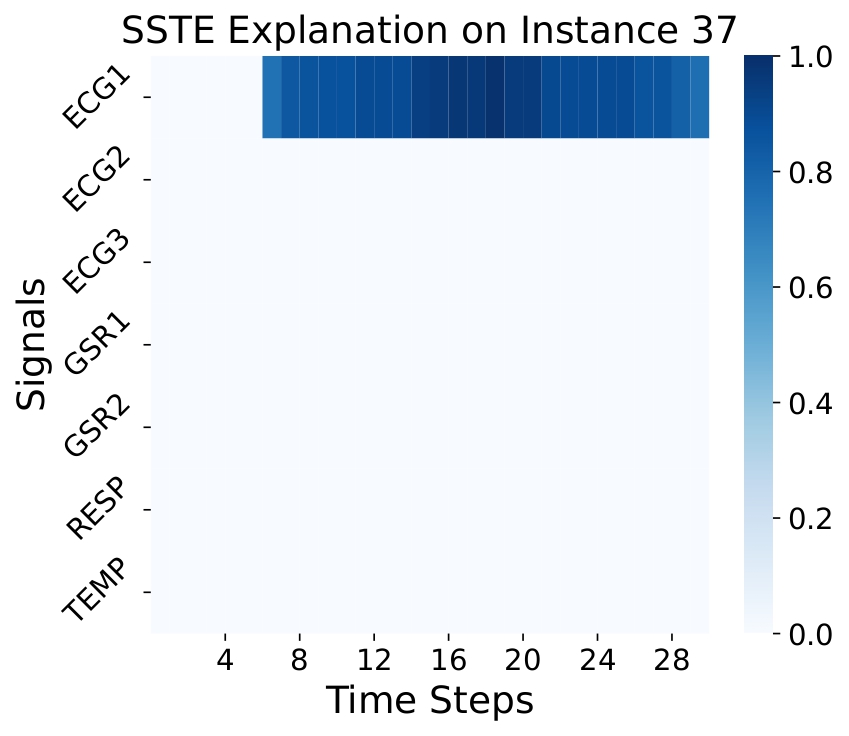}
   
    \caption{Dynamask (a), IG (b), LIME (c), and SSET (d) explanations of instance $37$ in MAHNOB-HCI\@. Heatmap visualizations present the explanations by mapping importance scores between 0 and 1 to a blue color spectrum. Like the results on WESAD (see Figs.~\ref{fig.explaination_WESAD} and~\ref{fig.explaination_dual_WESAD}), none of the XAI models except SSET gives a  meaningful explanation for the \textit{amusement} decision. While SSET shows the highest score ($0.9$) for the sub-sequence between 15 and 21, and $0.85$ for the rest of the sequence between 7 and 30, the explanations of other benchmarks do not make sense. The provided scores in the benchmarks are $0.5$ for Dynamask and almost $0$ for IG and LIME.}
    \label{fig.explaination_MAHNOB}
\end{figure*}

When applying the benchmarks and SSET to the MAHNOB-HCI dataset, the results confirm the previous findings on WESAD\@. The uninformative explanations of the benchmarks for instance $37$ can be seen in Fig.~\ref{fig.explaination_MAHNOB}, which shows that Dynamask cannot detect any variations in the mask and that LIME does not benefit from a promising nonlinear approximation of the black-box. The IG model shows poor attributions assigned to the ECG and GSR1 signals, while SSET can provide a salient sub-sequence of the ECG1 signal that contributes to the detection of the \textit{amusement} state. Given the manipulation of the sub-sequence, the importance at this state is above $0.9$ between steps 15 and 21 and is, on average, $0.85$ for the steps between 7--14 and 22--30.

\subsection{Salient Signals Evaluation} 
\label{sec.signals}
In this subsection, we look at how the importance is distributed across the signals in the $200$ test data. The goal is to understand which sensory data are mostly of the focus of CN-Waterfall to make decisions. The results can shed light on the solution to the problem of resource-limited laboratories so that the most efficient sensors can be employed on experiments.

\begin{figure*}[t]
    \centering
    \includegraphics[width=0.4\textwidth]{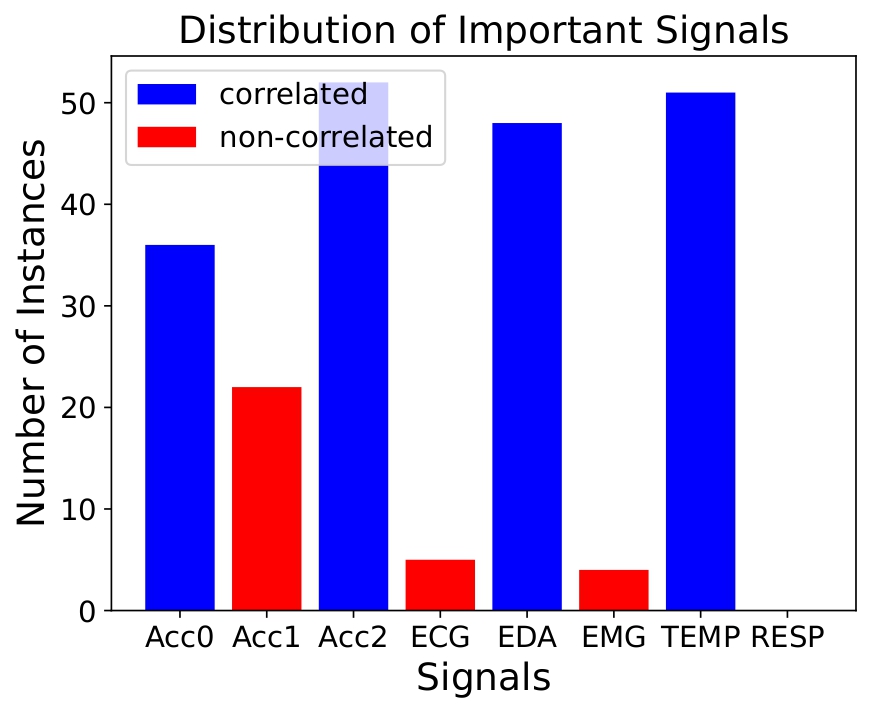}
    \includegraphics[width=0.4\textwidth]{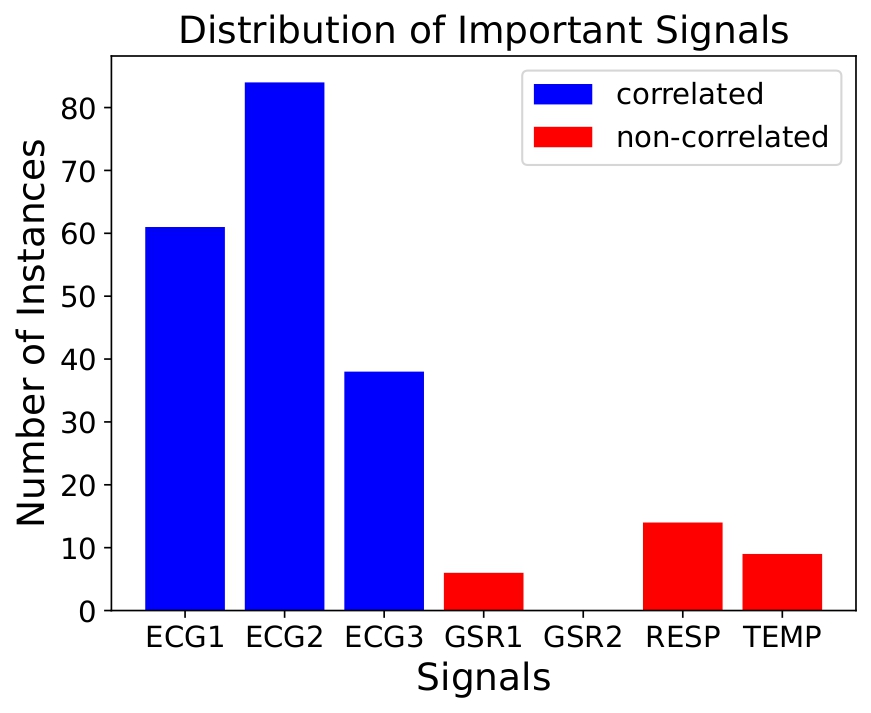} 
    \caption{Distribution of salient signals over a set of $200$ test data on WESAD (a) and MAHNOB-HCI (b). As shown, the correlated signals specified in~\cite{Fouladgar_CN:2020} contribute to the CN-Waterfall decision in most instances of both datasets.}
    \label{fig.signals}
\end{figure*}

The histogram of salient signals on WESAD is shown in Fig.~\ref{fig.signals}(a). As can be seen, CN-Waterfall has focused mainly on the ACC0, ACC2, EDA, and TEMP signals listed as correlated in~\cite{Fouladgar_CN:2020}. This finding reflects that the corresponding fusions in the CN-Waterfall structure play substantial roles compared to the non-correlated fusion in the intermediate-level data representation. The histogram of salient signals on MAHNOB-HCI (Fig.~\ref{fig.signals}(b)) also corroborates the attention of CN-Waterfall on the correlated signals. On MAHNOB-HCI, the ECG1, ECG2, and ECG3 signals demonstrate their salience in many more instances compared to the other signals. Therefore, one can conclude that solely applying the correlated signals is sufficient to make a decision in the resource-limited settings.  

\subsection{Window Size Evaluation} 
\label{sec.window_ize}
Here, we evaluate in what context the decided affective states are suppressed. In other words, the distribution of window sizes is measured over the $200$ test data to give us insight into the granularity of data sampling in terms of informativeness. 

\begin{figure*}[t]
    \centering
    \includegraphics[width=0.4\textwidth]{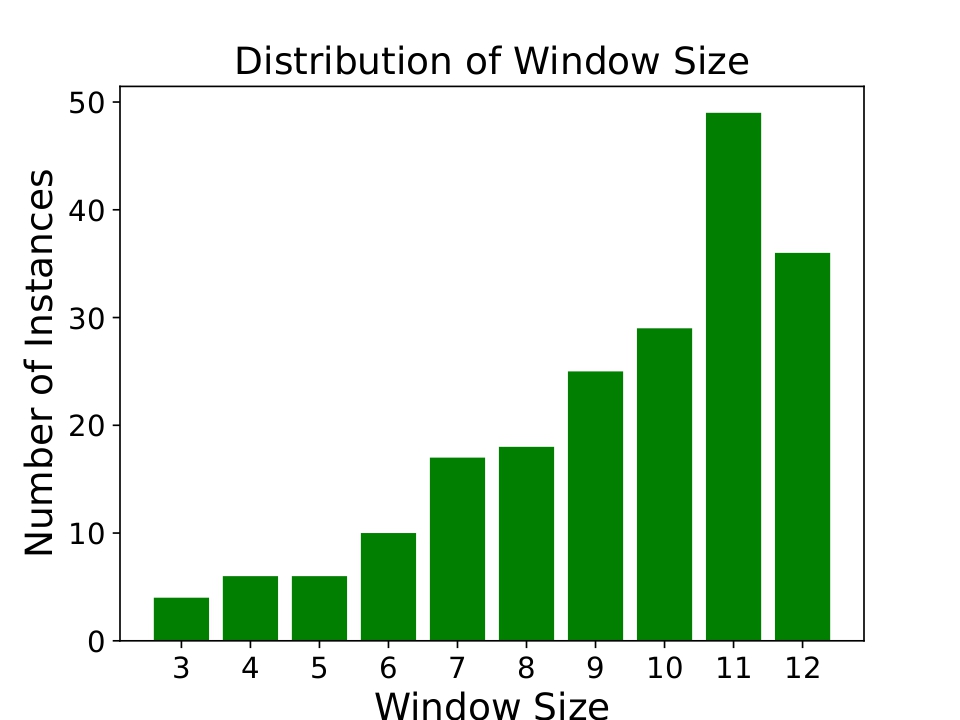}
    \includegraphics[width=0.4\textwidth]{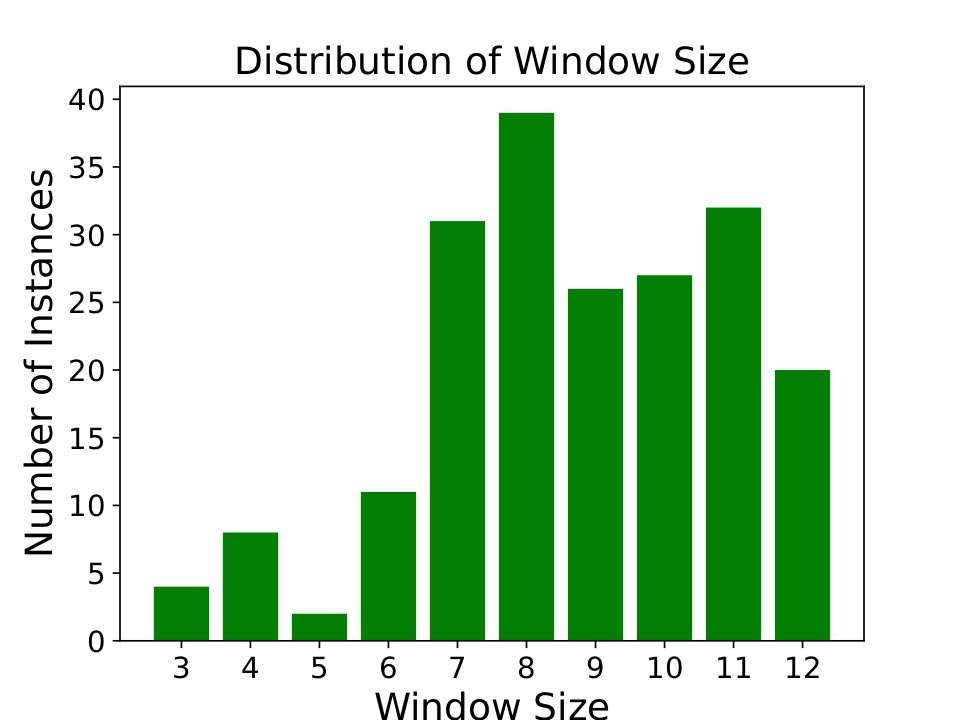}
    \caption{Distribution of window size over a set of $200$ test data on WESAD (a) and MAHNOB-HCI (b). The figure shows that the dynamic of states could be altered by salient sub-sequences of size $11$ and $8$ in most instances of WESAD and MAHNOB-HCI, respectively.}
    
    \label{fig.WindowSize}
\end{figure*}

Fig.~\ref{fig.WindowSize}(a) shows that on WESAD, the salient sub-sequences with a size of $11$ (more than $1$ second of data) reduce the prediction score at class $c$ in the majority of cases. On MAHNOB-HCI (Fig.~\ref{fig.WindowSize}(b)), the states of \textit{amusement}, \textit{happiness}, and \textit{surprised} could be altered within less than $1$ second. More precisely, the dynamic of most instances can be changed by sub-sequences of size $8$. 
We also confirm that there are highly informative signals in $5\%$ of the selected test data on both datasets with the size of salient sub-sequences less than or equal to $5$. However, we report an average of $9$ time steps that are important, in general.

\subsection{Window Size and Neighboring Distance Relation} 
\label{sec.correlation_ws_neighbors}
We investigate whether there is any relation between the distance of the sampled neighbors from the instance to be explained and the detected window size. To this end, we examine the scatter plot between these two factors and infer their correlation.

\begin{figure*}[ht]
    \centering
    \includegraphics[width=0.4\textwidth]{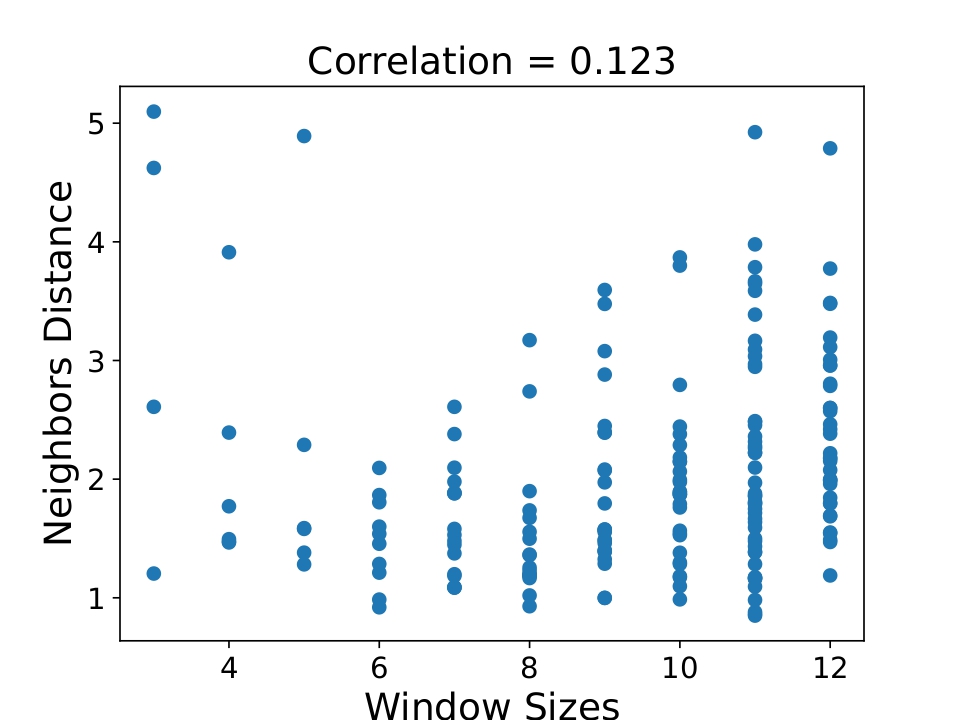} 
    \includegraphics[width=0.4\textwidth]{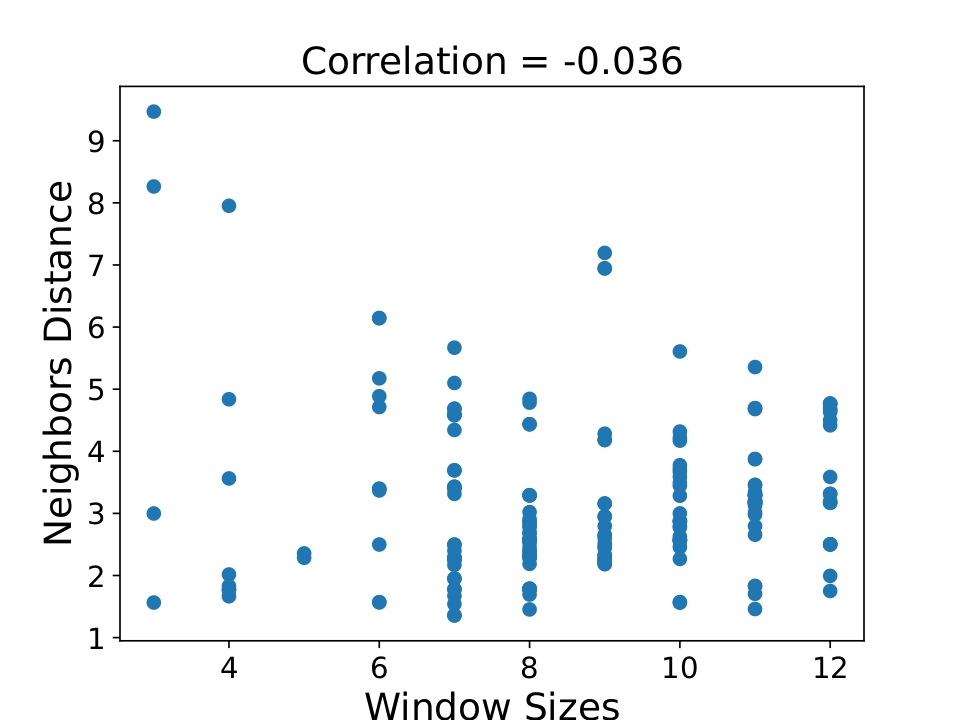} 
    \caption{The relation between the distance of neighboring samples from the instance to be explained and the window size on WESAD (a) and MAHNOB-HCI (b). One can infer that there is almost no correlation between the two factors on both datasets.}
    \label{fig.CorrWsNg}
\end{figure*}

As Fig.~\ref{fig.CorrWsNg}(a) illustrates, most sampled train data fall in the neighborhood with an average distance of less than $4$ on WESAD\@. There is a low correlation of $0.12$ between the distance and the size of salient sub-sequences, which can be ignored. In the second dataset (Fig.~\ref{fig.CorrWsNg}(b)), we see that most neighbors are located at an average distance of less than $6$ from the instance to be explained. Similar to WESAD, we report a negligible correlation between the latter distance and the window size ($-0.03$). It should be noted that the Pearson correlation is used in both dataset experiments.

\subsection{Explanation Quality of Explainers} 
\label{sec.Quality_explaination}
To gain an overall view, the quality of the explanations generated by the state-of-the-art XAI models and SSET are compared on both datasets. We explore the explanation quality measured in terms of \textit{precision}, \textit{informativeness}, \textit{similarity}. 

We measure the precision as the amount of suppression caused by the manipulated instances at the class of interest~\cite{Crabbe:2021}. A larger drop results in a higher precision of the explainer. We also postulate informativeness as the smallest sub-sequences that maximally drop the prediction score, similar to Fong's suggestion on images~\cite{Fong:2017}. As pointed
out by Miller~\cite{MILLER:2019}, the goal is to reduce the cognitive load of generated explanations for end users. Regarding similarity~\cite{Lang:2022}, close manipulated data to the instance to be explained are of interest to examine the black-box variations locally. Here, the Euclidean distance between the manipulated data and the instance of interest is calculated. Note that for the quality metrics, we take the average across the $200$ instances. 

As Tables~\ref{tab.XAIquality_WESAD} and~\ref{tab.XAIquality_MAHNOB} show, SSET outperforms the benchmarks in precision and informativeness on both datasets, at the expense of lower similarity. In essence, SSET provides the explanation with the precision of $0.1$ and $0.07$ on WESAD and MAHNOB-HCI, respectively, thanks to the neighborhood strategy and SSET components. It is worth mentioning that the model could perform even more precisely if the threshold $thr_c$ was set to a value lower than $0.5$. In contrast, LIME produces “blind” perturbations around the instance of interest, which result in insignificant prediction drops ($0.03$ on WESAD and $0.02$ on MAHNOB-HCI). In IG, the inherent dependency of explanations on the baseline and gradient-based characteristic impede crucial suppressions on the CN-Waterfall decision ($0.04$ and $0.03$ on WESAD and MAHNOB-HCI, respectively). Dynamask also strongly depends on the mask initialization to eventually build a perturbed instance that could reduce the black-box performance. However, a naive choice of mask ends up with unpromising perturbations, rejecting the performance drop. In our work, we employ mask values equal to $0.5$ as suggested by the original paper~\cite{Crabbe:2021}, resulting in no drop in the black-box output and zero precision.

On average, SSET discerns $9$ time steps as contributing to the detection of the current state in CN-Waterfall, while LIME approximates a linear model with respect to all time steps ($30$) of the instance variable. Similarly, IG and Dynamask take the gradients over $30$ steps of each variable. One could infer that more informative explanations are presented by SSET, even if the benchmarks generate valid explanations.  

Regarding the similarity measurement, SSET performs better than LIME by $2.02$ and $31.8$ on the first and second datasets, respectively. In LIME, undue distances between the perturbed data and the instance of interest 
make this model ineffective ($15.49$ on WESAD and $16.51$ on MAHNOB-HCI). Although the corresponding distances are lower in IG ($1.46$ on WESAD and $2.23$ on MAHNOB-HCI) and Dynamask ($0.39$ on WESAD and $0.57$ on MAHNOB-HCI), the latter models cannot provide valid explanations.


\begin{table}[ht]
  \centering
  \caption{Comparison of explanation quality between explainers on WESAD (best results shown in bold).}
  \label{tab.XAIquality_WESAD}
    \begin{tabular}{@{}llll@{}}
		\toprule
 	Explainer & Precision & Informativeness & Similarity \\
		\midrule
		LIME & 0.03 & 30 & 15.49  \\
		IG & 0.04  & 30 & 1.46 \\
		Dynamask & 0.00 & 30 & \textbf{0.39} \\
		SSET & \textbf{0.1} & \textbf{9} & 2.02 \\
		\botrule
	\end{tabular}
\end{table}

\begin{table}[ht]
  \caption{Comparison of explanation quality between explainers on MAHNOB-HCI (best results shown in bold).}\label{tab.XAIquality_MAHNOB}
  \begin{tabular}{@{}llll@{}}
  \toprule 
  Explainer & Precision & Informativeness & Similarity \\
  \midrule
  LIME & 0.02 & 30 & 16.51  \\
  IG & 0.03  & 30 & 2.23 \\
  Dynamask & 0.00 & 30 & \textbf{0.57}  \\
  SSET & \textbf{0.07} & \textbf{9} & 3.18  \\
\botrule
\end{tabular}
\end{table}


\section{Conclusion and Future Works}
\label{sec.Conclusion}
In this paper we introduced SSET, a post-hoc local XAI model applicable to time series in the affect detection domain. With SSET, we aimed to provide contextual explanations for the outcomes of non-differentiable detectors. To this end, two stages of swapping and sliding explored salient signals and intervals, respectively. Moreover, the degree of salience was measured in a novel fashion by incorporating multiple factors. We performed comprehensive experiments on CN-Waterfall, a highly accurate deep convolutional affect detector, and two datasets, WESAD and MAHNOB-HCI\@. Comparing SSET with LIME, IG, and Dynamask, we demonstrated the superiority of our proposed model over the benchmarks in terms of \textit{precision} and \textit{informativeness}. Furthermore, it was shown that the correlated signals contribute significantly to the decisions of CN-Waterfall. These results improve the transparency of ML models and assist experts in choosing the most appropriate sensors in resource-limited laboratory settings.

SSET was the only model that produced promising explanations. However, this model requires more computational time than the others, especially for the WESAD dataset. It took about $3$ hours to run for this dataset, while LIME and IG  took about $1$ minute and  Dynamask took about $1$ hour. With respect to the MAHNOB-HCI dataset, our proposed model and Dynamask ran for about $1$ hour to produce explanations, whereas LIME and IG ran for about $1$ minute.   

In the future, it should be possible to apply heuristics and efficient search algorithms targeting the neighborhood scope and context size to reduce the computation time. 
Another avenue would be to investigate SSET's quality on other time series and applications. It would be useful to examine different values of the dropping threshold to gain a better assessment of the precision. Finally, SSET could be studied in privacy-aware settings where time series should not be disclosed. Such study allows practitioners to understand deeper whether/to what extent SSET meet data protection measures. As scholars claim that explanability distorts privacy~\cite{grant_wischik:2020}, it is recommended to examine the proposed model under such settings before employing in the wild.

\section*{Declarations}

\begin{itemize}
\item \textbf{Fund:} This paper is funded by Ume\r{a} University
\item \textbf{Conflict of interest:} The authors declare that they have no conflict of interest. 
\item \textbf{Competing Interests:} No competing interests. 
\item \textbf{Data availability and access:} The WESAD dataset is publicly available in \url{https://www.archive.ics.uci.edu/dataset/465/wesad+wearable+stress+and+affect+detection} and the MAHNOB-HCI dataset is academically available via creating a request account in \url{https://mahnob-db.eu/hci-tagging/}.
\item \textbf{Authors contribution:} Conceptualization N.F., Implementations N.F., Experiments N.F., Analysis N.F. and K.F. and M.A., Writing original draft N.F., Reviewing N.F. and M.A., Supervision K.F. and M.A.
\item \textbf{Ethical and informed consent for data used:} We did not gather the data ourselves, instead we utilized the already collected WESAD and MAHNOB-HCI datasets for the outlined purposes of this paper. The authors in \cite{Schmidt:2018} had already published WESAD publicly and we obtained the dataset via the URL above. We also obtained MAHNOB-HCI by creating a request account and following the instructions provided in the above URL. Therefore, we adhered to the ethical guidelines and relevant regulations for the data used. 
\end{itemize}

\bibliography{sn-bibliography}

\end{document}